\pdfoutput=1

\documentclass[11pt]{article}

\usepackage{EMNLP2023}

\usepackage{times}
\usepackage{latexsym}

\usepackage[T1]{fontenc}

\usepackage[utf8]{inputenc}

\usepackage{microtype}

\usepackage{inconsolata}

\usepackage{graphicx}
\usepackage{booktabs}
\usepackage{multirow}
\usepackage{subcaption}
\usepackage{makecell}
\usepackage{float}

\usepackage{todonotes}

\title{Few-shot clinical entity recognition in English, French and Spanish: \\masked language models outperform generative model prompting}

\author{Marco Naguib, Xavier Tannier, Aurélie Névéol \\
  Université Paris-Saclay, CNRS, LISN, Orsay, France\\
  Sorbonne Université, Inserm, Université Sorbonne Paris-Nord, Limics, Paris, France\\
  \texttt{firstname.lastname@lisn.upsaclay.fr} \\
  \texttt{xavier.tannier@sorbonne-universite.fr} \\
  }

\begin{document}
\maketitle
\begin{abstract}


Large language models (LLMs) have become the preferred solution for many natural language processing tasks. In low-resource environments such as specialized domains, their few-shot capabilities are expected to deliver high performance. 
Named Entity Recognition (NER) is a critical task in information extraction that is not covered in recent LLM benchmarks. There is a need for better understanding the performance of LLMs for NER in a variety of settings including languages other than English. 
This study aims to evaluate generative LLMs, employed through prompt engineering, for few-shot clinical NER. 
We compare 13 auto-regressive models using prompting and 16 masked models using fine-tuning on 14 NER datasets covering English, French and Spanish. 
While prompt-based auto-regressive models achieve competitive F1 for general NER, they are outperformed within the clinical domain by lighter biLSTM-CRF taggers based on masked models. Additionally, masked models exhibit lower environmental impact compared to auto-regressive models.
Findings are consistent across the three languages studied, which suggests that LLM prompting is not yet suited for NER production in the clinical domain. 

\end{abstract}

\section{Introduction}
Electronic Health Records (EHR) are rich sources of clinical information \cite{fushman2009decision}, which often appear in unstructured text only \cite{escudie2017novel}. 
Efficiently extracting information from EHRs into a more structured form can help advance clinical research, public health surveillance and 
clinical decision support \cite{wang2018applications}.

Named Entity Recognition (NER) is a critical primary step in information extraction that aims to identify and categorize mentions of relevant entities in text. 
In the context of clinical information extraction, 
these can be mentions of clinical entities such as disorders or drugs. Extracting these entities can significantly enhance concept normalization \cite{cho2017normalization, wajsburt2021normalization, sung2022normalization} as well as facilitate interpreting patient profiling and phenotyping \cite{gerardin2022multilabel}.  
Clinical NER is widely regarded as a challenging problem : clinical entities are often jargon or ambiguous, and clinical texts have a nonstandard phrasal structure \cite{luo2020n2c2,leaman2015challenges}.
Additionally, the sensitive nature of EHRs results in a lack of publicly available clinical corpora, which are often restrictively licensed and predominantly available in English.
Moreover, the annotation of clinical NER data demands substantial domain expertise, rendering such campaigns both costly and time-intensive \cite{luo2020n2c2, neveol2014quaero, dogan2014ncbi, baez2020chilean}.
Additionally, due to the diversity
of clinical cases, data annotated for one biomedical application might not
necessarily be helpful for another.
Consequently, there is a critical need for data-efficient clinical NER, specifically few-shot NER approaches.

Large Language Models (LLMs) - specifically, causal, generative models - have demonstrated significant promise in few-shot learning across a wide variety of tasks, including text classification, machine translation, and question answering \cite{brown2020language, radford2019language}. This supports our main research question: Can few-shot learning performance of LLMs transfer to the task of named entity recognition?  

This study aims to evaluate generative LLMs, employed through prompt engineering, for few-shot clinical NER from the perspective of F1 performance and environmental impact.
\paragraph{Challenges of few-shot NER}
We identify methodological challenges with the evaluation of few-shot prompting for NER.

First, there is no standard, widely adopted manner of prompting LLMs for NER tasks \cite{li2022surveyner, shen2023locating, wang2023gptner, keraghel2024survey}, resulting in significant challenges for reproducibility and variations in results that are difficult to interpret.
Second, many efforts towards "few-shot learning" with LLMs design prompts based on their performance on large held-out validation datasets \cite{brown2020language, tam2021improving, radford2021learning, qin2021learning}, which is not consistent with a few-shot set-up.
In addition, in-context learning performance is shown to depend greatly on the prompt structure: a small change in task phrasing, the examples presented, the order of examples, or the tagging format can affect the performance \cite{zhao2021calibrate,lu2022fantastically, min2022rethinking}. Therefore, making these choices assuming large annotated validation dataset leads to performances that are shown \cite{perez2021true} to be over-optimistic and impossible to find in a real few-shot setting.
Third, \citet{zaghir2024prompt} show that the majority of recent studies employing prompt engineering in medical applications lack a non-prompt-related baseline, such as fine-tuned BERT-like Masked Language Models (MLMs), which complicates the accurate assessment of the performance of these LLMs.
Finally, most of these studies are mainly concentrated on English, and based on GPT, which is mainly trained on English, \cite{jimenez2022thinking, wang2023gptner, ashok2023promptner, hu2023zero, zaghir2024prompt}, limiting the generalizability of evaluations to other languages.
The contributions of this work are as follows:
\begin{enumerate}
    \item We describe a systematic algorithm for creating and optimizing prompts for NER, bringing a particular attention to tagging prompts \cite{wang2023gptner}, a novel NER prompting technique that recently showed particular promise \cite{garcia2024gpt, magron2024jobskape}.
    \item We evaluate our algorithm in a true few-shot setting, by allowing prompt optimization only on the few annotated instances through cross-validation. 14 NER tasks were evaluated, spanning 6 general-domain datasets and 8 clinical datasets, with a focus on English, French and Spanish.
    \item We compare this approach, applied to 13 generative LLMs, to the standard fine-tuning approach, applied to 16 MLMs, both in terms of performance and environmental impact. 
    We  provide our code at \mbox{\small{\url{github.com/marconaguib/autoregressive_ner}}}
\end{enumerate}

\section{Few-shot \& clinical NER}
\label{background}

\paragraph{Few-shot NER with pre-trained MLMs}
\label{background-MLM}

Leveraging MLMs for NER usually involves using them as encoders, and training a linear projection to map vector representations into an NER tagging of the sentence, while jointly fine-tuning the parameters of the language model itself for the downstream task of NER \cite{devlin2019bert}.
This approach has been widely studied \cite{liu2019finetune, petroni2020context, joshi2020spanbert,schweter2021flert}.
\citet{wajsburt2021these} propose a similar architecture, enhanced with an entity decoder, to iteratively predict entity spans in the input, allowing the model to detect nested entities.

Few-shot NER can be performed by simply training such systems with the limited data available. Other approaches have been proposed to leverage MLMs specifically in few-shot setting.
Namely, metric learning \cite{fritzler2019few, yang2020simple, huang2021few} proposes to train systems to instead learn a metric over the output space. New instances can then be classified based on the distance separating them from other labeled instances. Label encoding \cite{aly2021leveraging, ma2022label, hou2020few} suggests, instead, to leverage label names or textual label descriptions and encode them along with the instances in order to better tag them.

\paragraph{Few-shot NER with generative LLMs}
Recently, prompt construction has gained interest in the community \cite{brown2020language,liu2023pre}. While most related work focused on studying prompt formulation and exploring better-performing prompt structures \cite{wei2022chain, ashok2023promptner, vilar2023prompting, wang2023gptner} also known as "prompt engineering", other work proposed continuous optimization of the prompt through prompt tuning \cite{ma2022templatefree, layegh2023contrastner, hu2023vpn}, usually reporting marginal improvements over baselines.

There is no standard, widely adopted manner of building NER prompts \cite{liu2023pre}. In fact, NER associates to each instance a set of spans, each of which having a type. This structured nature of the prediction make it hard to find an intuitive but efficient manner to prompt a language model for NER, that adapts well to all contexts.

For instance, the main practice is to use separate prompts for different entity types \cite{li2020unified,liu2022qaner, chen2023learning}. This choice seems well-suited when the task is interested in a handful of types of entities (typically 5-10).
When interested in less entity types, a single prompt can be used for detecting all entities \cite{ashok2023promptner}.
On the other hand, if there is more entity types, it could be interesting to enumerate every possible span in the input sentence and let the model predict the entity type of the span, if any \cite{cui2021template}. This method, on the inverse, is impractical for long inputs.

We identify three strategies for prompting LLMs.
Constrained prompting attempts to better formulate the NER task by constraining the generation to fill in specific hand-crafted templates, usually adapted to MLMs \cite{cui2021template, shen2023locating,ye2023decomposed, schick2021just}.
Listing prompts consist in simply making the language model predict the entities in a list \cite{ashok2023promptner}.
Tagging prompts were studied more recently by \citet{wang2023gptner}. They make the language model surround entity mentions with special tags.


\paragraph{Few-shot clinical NER}
MLM-based few-shot NER has also been explored in the biomedical domain \cite{geJBI2023}. Metric leaning \cite{yang2020simple} and label encoding \cite{aly2021leveraging, ma2022label} have been explored, as well as other approaches such as active learning \cite{kormilitzin2021med7}, supervised pretraining \cite{huang2021baseline} and prompt-based learning \cite{lee2022good, chen2023few, cui2024language}.

Few studies have focused on LLM-prompting-based few-shot clinical NER. In \citet{HuJAMIA2024}, GPT-3 and ChatGPT are evaluated on the 2010 i2b2/VA task \cite{uzuner20112010} in a zero-shot context. In \citet{jimenez2022thinking}, GPT-3 is evaluated on a set of biomedical information extractions tasks including the NCBI-Disease \citet{dogan2014ncbi}. 

On languages other than English,\citet{meoni2023large} use InstructGPT-3 is used to build multilingual training corpora to train smaller models, and in \citet{ateia2023chatgpt}, ChatGPT is evaluated on an NER challenge focused on extracting medical procedures in Spanish.

Another interesting direction is partly fine-tuning  \cite{liao2023parameter} a general-domain LLM on clinical text \cite{han2023medalpaca, toma2023clinical, yang2024enhancing}, and prompting the resulting LLM.

\section{Named Entity Recognition Experiments}

\subsection{Evaluation tasks}
We use 14 publicly-available NER datasets (described in the next section) to compare prompted causal models to fine-tuned masked language models in few shot settings.  
For each study language, we selected two out-domain datasets and two or three in-domain datasets, aiming to use comparable resources (same genre, tagset, annotation guidelines) across languages whenever possible. We use official training, validation, and test subsets when available; otherwise, we apply an 80\%-10\%-10\% split for training, validation, and testing.
\subsubsection{General-domain evaluation datasets}
\textbf{WikiNER} \cite{nothman2013wikiner} is a multilingual \textit{silver-standard} annotated NER dataset. It consists of a late-2010 snapshot of Wikipedia in nine languages. Hyperlinks referring to persons, locations or organizations were automatically annotated. We use the English, French and Spanish versions of this dataset.

\textbf{CoNLL-2002} \cite{sang2002conll} and \textbf{CoNLL-2003} \cite{sang2003conll} are two manually-annotated multilingual NER dataset released as a part of CoNLL shared tasks. Mentions of persons, locations, organizations and miscellaneous entities are annotated. We use the Spanish data of the 2002 version, which is a collection of news wire articles made available by the Spanish EFE News Agency, released in May 2000. We use the English data of the 2003 version, which consists of Reuters news stories between 1996 and 1997. 

\textbf{Quaero French Press} \cite{grouin-etal-2011-proposal} is a manually annotated corpus of about 100 hours of speech transcribed from French speaking radio broadcast. This corpus was used in the 2011 Quaero named entity evaluation campaign. It comprises annotations for 5 entity types further divided into 32 subtypes. Our experiments relied on the five entity types: persons, locations, organizations, functions, and facilities.

\subsubsection{Clinical evaluation datasets}

\textbf{E3C} \cite{magnini2021e3c} is a European multilingual corpus (Italian, English, French, Spanish, and Basque) of semantically annotated clinical narratives. The texts are collected from multiple publicly-available sources such as abstracts extracted from CC-licensed journals.
We use the gold standard material available from the English, French and Spanish versions of this dataset. The clinical narratives are annotated with 6 entity types : actors, body parts, events, RMLs (measurements and test results) and clinical entities. 

The \textbf{n2c2-2019} \cite{luo2020n2c2} shared task focuses on medical concept normalization. It uses the MCN corpus developed by \cite{luo2019mcn}, often referred to as the n2c2-2019 dataset. It includes de-identified discharge
summaries from the Partners HealthCare and Beth Israel Deaconess Medical Center. In order to convert the medical concept normalization task into an NER task, we use the annotated Concept Unique Identifiers (CUIs) to map each mention to the corresponding UMLS semantic group \cite{lindberg1993umls,mccray2001aggregating}.

The \textbf{NCBI-Disease} \cite{dogan2014ncbi} corpus gathers 793 PubMed abstracts where mentions of diseases are annotated in four types depending on their syntax : Specific Diseases (e.g. \textit{diastrophic dysplasia}), Disease Classes (e.g. \textit{an  autosomal recessive disease}), Composite Mentions (e.g. \textit{colorectal, endometrial, and ovarian cancers}),  and Modifiers (e.g. \textit{\textit{C7-deficient}}) .

\textbf{QuaeroFrenchMed} \cite{neveol2014quaero}  consists of two text sources that we treat separately. The first part, \textbf{EMEA} is a collection of 13 patient information leaflets on marketed drugs from the European Medicines Agency (EMEA). The second part, \textbf{MEDLINE}, consist of 2,500 titles of research articles indexed in the MEDLINE database\footnote{\label{note_pubmed}\url{http://pubmed.ncbi.nlm.nih.gov/}}. The two parts are annotated with 10 entity types corresponding to UMLS semantic groups. 

\textbf{The Chilean Waiting List} \cite{baez2020chilean} corpus consists of 900 de-identified referrals for several specialty consultations in Spanish from the waiting list in Chilean public hospitals, manually annotated with 10 entity types : abbreviations, body parts, clinical findings, diagnostic procedure, diseases, family members, laboratory or test results, laboratory procedures, medications, procedures, signs or symptoms and therapeutic procedures. It can be noted that these types can be redundant (e.g. all diagnostic procedures are also annotated as procedures).

\subsubsection{Few-shot learning set-up}

We simulate the few-shot context by only 
providing the models with
a few annotated examples that can be used in training, prompting and validation. 
No additional examples are made available. 
In this study, we choose to mainly focus on $k=100$ sentences, which corresponds to one to two hours of annotation in the clinical domain \cite{neveol2014quaero, campillos2018french}.
We use a fixed random seed $p$ to choose $k$ examples among all those available in the 
corpus.
In Section~\ref{ablation-sample}, we discuss the effect of the choice of $k$ and of $p$.

Additionally, we train the best-performing language models with the entire training dataset to provide a skyline comparison, i.e. performance of the models outside the few-shot setting.

\subsection{Language Models}

Table \ref{tab:LM_features} (appendix \ref{appendices-models}) presents an overview of the language models used in our study. English is covered in all causal language models, which is not the case for Spanish and French. Except for mBERT and XLM-RoBERTa, masked language models cover only one of our study languages.

\subsection{NER with MLMs}

Although MLMs have been adapted to few-shot learning in architectures suited for low-resource contexts (see section~\ref{background-MLM}), we compare LLM prompting to the simple and standard MLM usage  without any further adaptation for few-shot learning, as an accessible and easily-reproducible baseline.

We use NLStruct \cite{wajsburt2021these}, an open-source Python library \footnote{\url{https://github.com/percevalw/nlstruct}} that implements the standard fine-tuning approach. NLStruct uses the representations provided by the language model to encode the input, then employs a bidirectional LSTM decoder and a CRF to iteratively predict the entities present in the encoded input, as described by \citet{gerardin2022multilabel}. This approach allows NLStruct to effectively handle nested entities, which are prevalent in some of the study corpora.

We train the model for 20 epochs on 80\% of the data and use the remaining held-out 20\% for early stopping.
\subsection{NER with generative LLM prompting}
\begin{figure*}[ht]
    \centering
    \includegraphics[width=\linewidth]{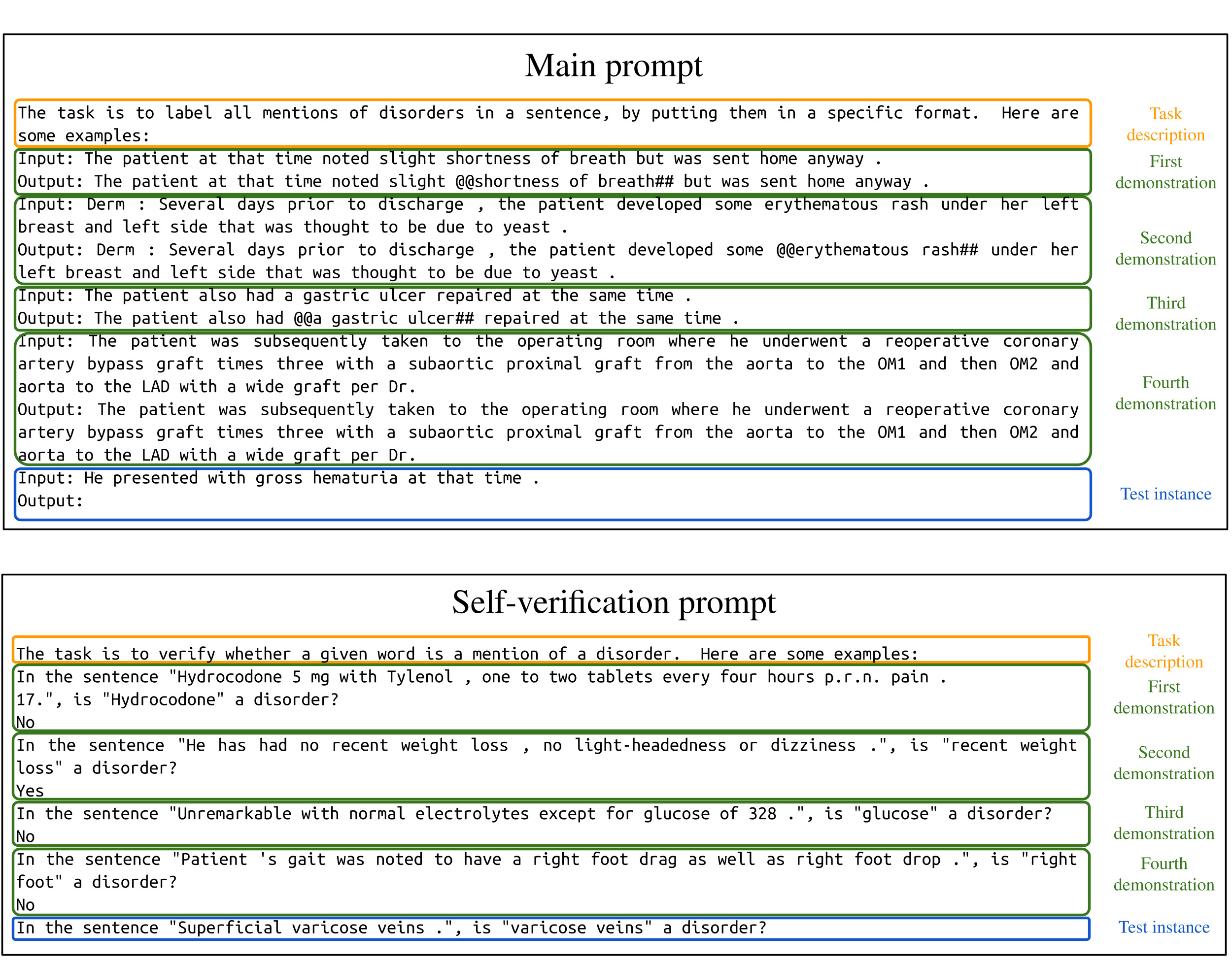}
    \caption{Example of a tagging prompt, used in the main experiment (top) and a self-verification prompt (bottom) for detecting DISO mentions in \textbf{n2c2-2019}}
    \label{fig:prompt}
\end{figure*}

Our experiments prompt models to tag entities in the input sentence, instead of listing them. This choice is supported by further experiments reported in Section~\ref{ablation-listing}. 
The upper part of Figure~\ref{fig:prompt} shows a sample tagging prompt, highlighting sections in the prompt that 
guided the design of features for prompt phrasing.

\paragraph{Prompt features} We describe below the nine optional features that control the phrasing of the prompt, as well as the criteria for selecting the few-shot examples featured in the prompt. 

\begin{enumerate}
    \item \textbf{Prompt language}: By default, we prompt all language models in English, as it is the most ubiquitous language in all of their training corpora. This feature allows the model to be prompted in French or Spanish, to align the prompt language with  that of the test sentence.
    \item \textbf{Additional sentences}: By default, we present 5 annotated sentences in the prompts. This feature presents 5 additional sentences (i.e., 10 sentences in total). Section~\ref{ablation-number} discusses adding more demonstations to the prompt.
    \item \textbf{Self verification}: By default, we select the 5 closest sentences to the test sentence in terms of TF-IDF distance, among the training set. The mentions tagged by the model are then considered to be the model's final predictions.
    This feature selects instead the 5 sentences featuring the most entities of the target type and features them in an initial prompt. Intuitively, this prompt results in higher recall and lower precision.  
    A second "self-verification" prompt is then used over the model's initial predictions in order to filter out the false positives.
    A sample self-verification prompt is shown in the bottom part of Figure~\ref{fig:prompt}.

    The number of demonstrations follows that of the main prompt.
    \item \textbf{Taggers}: By default, we follow \cite{wang2023gptner} prompting the model to surround mentions with \textit{@@} and \textit{\#\#}. This feature prompts it to surround mentions with quotes \textit{<<} and \textit{>>} instead.
    \item \textbf{Address a specialist in the prompt}: By default, the first sentence is the task description shown in Figure~\ref{fig:prompt}. This feature starts the prompt with \textit{You are an excellent <specialist>. You can identify all the mentions of <entity-type> in a sentence, by putting them in a specific format. Here are some examples you can handle:} instead. The \textit{<specialist>} is a \textit{linguist} or a \textit{clinician}, following the task domain.
    \item \textbf{Include label definitions in the prompt}: This feature adds a one-sentence description for each entity type. Full entity descriptions used can be found in appendix \ref{appendices-descriptions}.
    \item \textbf{Introductory sentence for the test instance}: By default, the demonstrations are immediately followed by the test instance. This feature separates them with \textit{Identify all the mentions of <entity-type> in the following sentence, by putting <begin-tag> in front and a <end-tag> behind each of them.}
    \item \textbf{Require a long answer for the self-verification}: By default, the self-verification prompt demonstrates \textit{Yes} (respectively \textit{No}) as answers. This feature demonstrates \textit{<mention> is a(n) <entity-type>, yes.} (respectively \textit{<mention> is not a(n) <entity-type>, no.}) instead.
    \item \textbf{Dialogue template}: This feature replaces the \textit{Input:} and \textit{Output:} in the prompt by dashes to imitate a dialog template.
\end{enumerate}
\paragraph{Identification of optimal prompt configuration} In-context learning performance is shown to vary greatly depending on the exact phrasing of prompts \cite{lu2022fantastically, min2022rethinking}. In addition, the optimal choice for each of these features can vary depending on the model used. For instance, intuitively, models that are heavily pretrained on the English language tend to perform better with an English template than one in the language of the corpus.

While our system aims to search for the best combination of parameters for each model, a grid search over them would require $2^9 = 512$ experiments for each model, for each dataset. In order to build a lighter system, we 
opt for
a greedy search. We iterate over the features 
and select options that perform better than the default. In Section~\ref{ablation-grid}, we illustrate how the optimal parameter combination can be obtained with a greedy search using 5\% of the computation required for grid search. 
Many efforts towards "few-shot learning" with LLMs design prompts on large held-out validation datasets \cite{brown2020language, tam2021improving, radford2021learning, qin2021learning}. This leads to results that are shown \cite{perez2021true} to be over-optimistic. A true few-shot evaluation involves optimization with access to a small number of annotated instances, which corresponds to our $k=100$. In this no-training context, we follow \cite{perez2021true} optimizing these features through a leave-one-out cross-validation (LOOCV) over the validation set.

\subsection{Performance metrics}
\paragraph{Micro-F1} For simplicity, we evaluate models over one global information extraction performance score. It is computed as the micro-average of F1-measures for each entity type.
\paragraph{Carbon footprint} We use GreenAlgorithms v2.2 \cite{lannelongue2021green} \footnote{http://calculator.green-algorithms.org/} to estimate the carbon footprint of each experiment, based on factors such as runtime, computing hardware and location where electricity used by our computer facility was produced.

\section{Results}

\subsection{Environmental Impact}

\begin{figure}
    \centering
    \includegraphics[width=0.9\linewidth]{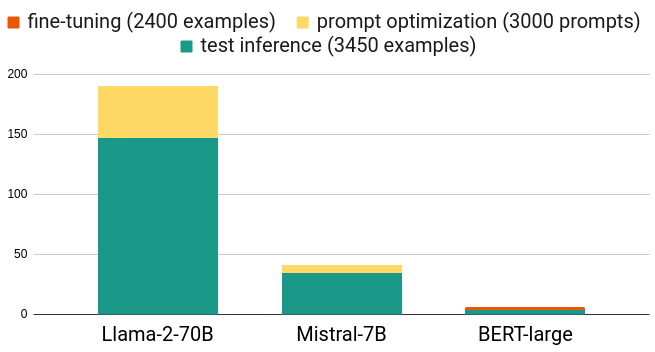}
    \caption{Carbon emission (g) incurred by resolving ConLL-2003 using three models}
    \label{fig:carbon}
\end{figure}

Figure~\ref{fig:carbon} compares carbon emission
incurred by resolving ConLL-2003 using LLM prompting (using Llama-2-70B and Mistral-7B) as well as with fine-tuning BERT-large. For comparison, the impact of using Llama-2-70B is comparable to that of driving a thermal car of average size for 1 kilometer.

Appendix \ref{appendices-carbon} details the carbon emission estimations for all of our experiments.
In total, the experiments described in this paper are estimated to have generated around 31kg of CO2 equivalent (29kg for main experiments, and 2kg for ablation).

\begin{table*}[ht]
\centering
\centerline{\scalebox{0.6
}{\begin{tabular}{lll|ccccc|ccccc|cccc}
 & & & \multicolumn{5}{c|}{English} & \multicolumn{5}{c|}{French} & \multicolumn{4}{c}{Spanish} \\
\cmidrule{3-7} \cmidrule{8-12} \cmidrule{13-16}
& \# & Model & WikiNER & CoNLL2003 & E3C & n2c2 & NCBI & WikiNER & QFP & E3C & EMEA & MEDLINE & WikiNER & CoNLL2002 & E3C & CWL \\
\midrule
\midrule
 \multicolumn{17}{l}{\textit{Few-shot approaches}} \\
\midrule
\multirow{13}{*}{\rotatebox[origin=c]{90}{Causal}} & 1 & Llama-2-70B & 0.728 & 0.721 & 0.312 & \textbf{0.309} & \textbf{0.400} & 0.740 & 0.400 & 0.483 & 0.201 & 0.312 & \textbf{0.805} & 0.616 & 0.021 & 0.339 \\
& 2 & Llama-3-8B-Instruct & \textbf{0.756} & \textbf{0.727} & \textbf{0.478} & 0.252 & 0.390 & \textbf{0.796} & \textbf{0.508} & \textbf{0.652} & \textbf{0.443} & \textbf{0.433} & 0.784 & \textbf{0.764} & \textbf{0.226} & \textbf{0.437} \\
& 3 & Mistral-7B & 0.754 & 0.646 & 0.488 & 0.291 & 0.395 & 0.727 & 0.428 & 0.590 & 0.229 & 0.333 & 0.720 & 0.707 & 0.083 & 0.374 \\
& 4 & Phi-3-medium-instruct & 0.464 & 0.302 & 0.234 & 0.112 & 0.178 & 0.394 & 0.322 & 0.317 & 0.139 & 0.161 & 0.612 & 0.461 & 0.045 & 0.280 \\
& 5 & BLOOM-7B1 & 0.524 & 0.557 & 0.279 & 0.113 & 0.151 & 0.148 & 0.206 & 0.320 & 0.197 & 0.120 & 0.470 & 0.419 & 0.051 & 0.117 \\
& 6 & Falcon-40B & 0.686 & 0.708 & 0.280 & 0.279 & 0.305 & 0.662 & 0.456 & 0.378 & 0.279 & 0.283 & 0.720 & 0.543 & 0.072 & 0.267 \\
& 7 & GPT-J-6B & 0.521 & 0.493 & 0.167 & 0.179 & 0.238 & 0.423 & 0.244 & 0.334 & 0.080 & 0.177 & 0.005 & 0.142 & 0.021 & 0.162 \\
& 8 & OPT-66B & 0.608 & 0.495 & 0.227 & 0.157 & 0.234 & 0.624 & 0.406 & 0.019 & 0.206 & 0.283 & 0.166 & 0.273 & 0.043 & 0.204 \\
& 9 & Vicuna-13B & 0.657 & 0.708 & 0.355 & 0.236 & 0.300 & 0.677 & 0.350 & 0.399 & 0.207 & 0.326 & 0.744 & 0.250 & 0.040 & 0.213 \\
& 10 & Vicuna-7B & 0.594 & 0.489 & 0.259 & 0.147 & 0.172 & 0.591 & 0.277 & 0.439 & 0.152 & 0.296 & 0.659 & 0.569 & 0.042 & 0.151 \\
& 11 & BioMistral-7B & 0.414 & 0.354 & 0.175 & 0.086 & 0.257 & 0.547 & 0.299 & 0.350 & 0.236 & 0.186 & 0.578 & 0.540 & 0.014 & 0.226 \\
& 12 & Medalpaca-7B & 0.537 & 0.586 & 0.272 & 0.138 & 0.132 & 0.529 & 0.142 & 0.259 & 0.162 & 0.252 & 0.581 & 0.490 & 0.088 & 0.220 \\
& 13 & Vigogne-13B & 0.593 & 0.655 & 0.252 & 0.176 & 0.309 & 0.515 & 0.250 & 0.464 & 0.099 & 0.142 & 0.580 & 0.561 & 0.010 & 0.198 \\
\midrule
\multirow{16}{*}{\rotatebox[origin=c]{90}{Masked}} & 14 & mBERT & 0.768 & 0.804 & 0.624 & 0.378 & 0.401 & 0.801 & 0.728 & 0.741 & 0.588 & 0.428 & \textbf{0.812} & 0.760 & 0.324 & 0.432 \\
& 15 & XLM-R-large & 0.786 & 0.826 & \textbf{0.637} & 0.462 & 0.471 & 0.811 & 0.781 & 0.762 & 0.629 & 0.531 & 0.797 & 0.781 & 0.325 & 0.528 \\
& 16 & BERT-large & 0.776 & 0.814 & 0.626 & 0.435 & 0.422 & - & - & - & - & - & - & - & - & - \\
& 17 & RoBERTa-large & \textbf{0.790} & \textbf{0.829} & 0.626 & 0.462 & \textbf{0.552} & - & - & - & - & - & - & - & - & - \\
& 18 & Bio\_ClinicalBERT & 0.528 & 0.542 & 0.621 & 0.469 & 0.420 & - & - & - & - & - & - & - & - & - \\
& 19 & ClinicalBERT & 0.462 & 0.597 & 0.622 & \textbf{0.480} & 0.397 & - & - & - & - & - & - & - & - & - \\
& 20 & MedBERT & 0.613 & 0.673 & 0.607 & 0.478 & 0.504 & - & - & - & - & - & - & - & - & - \\
& 21 & CamemBERT-large & - & - & - & - & - & \textbf{0.829} & \textbf{0.793} & 0.768 & \textbf{0.661} & \textbf{0.564} & - & - & - & - \\
& 22 & FlauBERT-large & - & - & - & - & - & 0.826 & 0.778 & 0.760 & 0.635 & 0.540 & - & - & - & - \\
& 23 & DrBERT-4GB & - & - & - & - & - & 0.587 & 0.599 & 0.730 & 0.602 & 0.497 & - & - & - & - \\
& 24 & CamemBERT-bio & - & - & - & - & - & 0.782 & 0.761 & \textbf{0.779} & 0.636 & 0.557 & - & - & - & - \\
& 25 & BETO & - & - & - & - & - & - & - & - & - & - & 0.794 & 0.732 & 0.352 & 0.522 \\
& 26 & PatanaBERT & - & - & - & - & - & - & - & - & - & - & 0.802 & 0.769 & 0.343 & 0.487 \\
& 27 & TulioBERT & - & - & - & - & - & - & - & - & - & - & 0.804 & \textbf{0.798} & 0.340 & 0.482 \\
& 28 & BSC-BioEHR & - & - & - & - & - & - & - & - & - & - & 0.804 & 0.758 & 0.354 & \textbf{0.578} \\
& 29 & BSC-Bio & - & - & - & - & - & - & - & - & - & - & 0.804 & 0.775 & \textbf{0.358} & 0.552 \\
\midrule
\midrule
\multicolumn{17}{l}{\textit{Masked fully-supervised (skyline)}} \\
\midrule
 & & RoBERTa-large & 0.919 & 0.939 & 0.718 & 0.712 & 0.815 & - & - & - & - & - & - & - & - & - \\
 & & CamemBERT-large & - & - & - & - & - & 0.928 & 0.834 & 0.828 & 0.748 & 0.713 & - & - & - & - \\
 & & BETO & - & - & - & - & - & - & - & - & - & - & 0.918 & 0.881 & 0.411 & 0.736 \\
\bottomrule
\end{tabular}}}
\caption{This table presents the micro-F1 obtained from few-shot experiments. Skyline results are obtained using all training data available instead of the few-shot setting.}
\label{tab:results}
\end{table*}



\subsection{Comparison of model performance}

\begin{figure}[!ht]
        \centering
        \includegraphics[width=\linewidth]{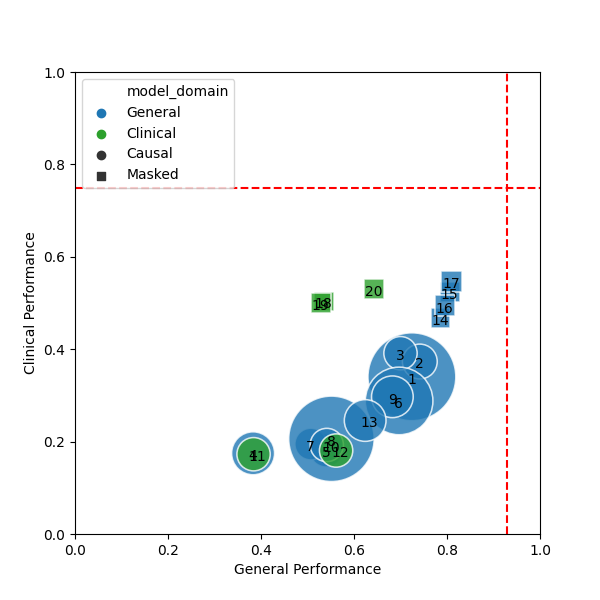}  
        \caption{Performance of models on English. The general performance is the average of micro-F1 obtained on WikiNER-en and CoNLL-2003. The clinical performance is the average on E3C-en, n2c2 and NCBI-Disease. The red lines represent the \emph{skyline} performance obtained with the entirety of each training dataset.}
        \label{fig:res-en}
\end{figure}

\begin{figure}[!ht]
        \centering
        \includegraphics[width=\linewidth]{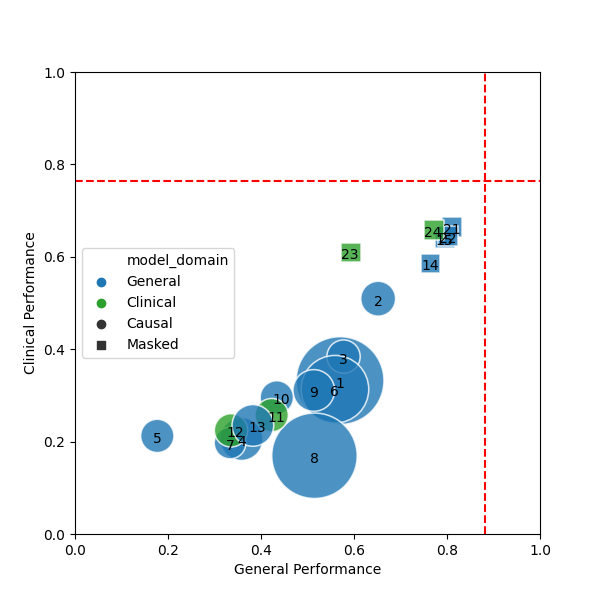}
        \caption{Performance of models on French. The general performance is the average of micro-F1 obtained on WikiNER-fr and QuaeroFrenchPress. The clinical performance is the average on E3C-fr, EMEA and MEDLINE. The red lines represent the \emph{skyline} performance obtained with the entirety of each training dataset.}
        \label{fig:res-fr}
\end{figure}

\begin{figure}[!ht]
        \centering
        \includegraphics[width=\linewidth]{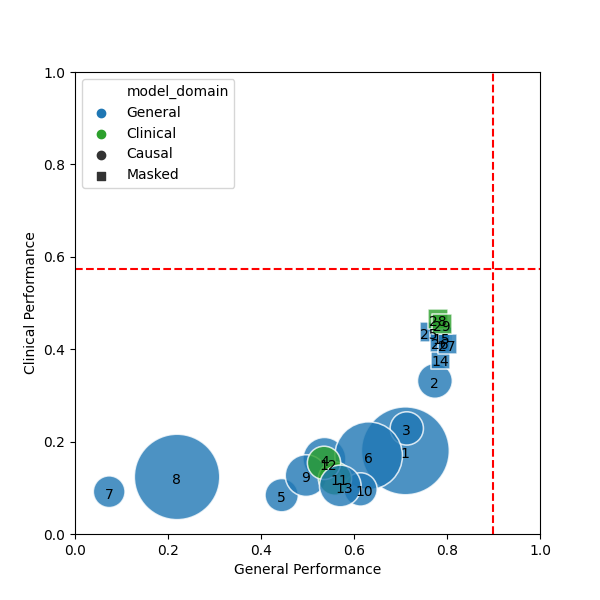} 
        \caption{Performance of models on Spanish. The general performance is the average of micro-F1 obtained on WikiNER-es and CoNLL-2002. The clinical performance is the average on E3C-es and CWL. The red lines represent the \emph{skyline} performance obtained with the entirety of each training dataset.}
        \label{fig:res-es}
\end{figure}

Table~\ref{tab:results} presents the micro-F1 performance of the models on the test set of each dataset. Figures~\ref{fig:res-en} to \ref{fig:res-es} present the results for each language with a vizualization of model type (circles for causal models vs. squares for masked models), model size in terms of parameters and domain (clinical in green vs. general in blue). 
Overall, results 
show that, despite being smaller and theoretically requiring a larger amount of training data, masked, "BERT-like" models consistently outperform generative LLM prompting in the context of few-shot NER
commonly found in biomedical applications.
Specifically, on English and Spanish, Figures~\ref{fig:res-en} and \ref{fig:res-es}, show that, while some generative LLMs are competitive in the general domain, they suffer a sharper performance drop in clinical NER, compared to fine-tuned MLMs. On French, Figure~\ref{fig:res-fr} suggests that generative LLM prompting does not seem competitive on either domain.

Moreover, this performance comes at a much lower environmental impact (CO2 emissions are 10-50 times lower for MLMs vs. prompted LLMs).
Another important finding is that, in addition to high performance scores, MLMs achieve results that are relatively close to each other. For example, on the WikiNER generalist task in English, the 4 general-domain models tested achieved F1-scores of between 0.768 and 0.79.


Besides, we show that
domain-adapted MLMs
(e.g., ClinicalBERT, CamemBERT-bio)
exhibit a sharp performance drop in general domain tasks, illustrating the classical issue of "catastrophic forgetting". 
In addition, they do not contribute
performance improvements in specialized tasks, 
with the exception of Spanish tasks. 
However, there is a notable difference in model size: specialized models only have 110 million parameters (vs. 340 million for other models).

Named entity recognition based on BERT-type representations has received a great deal of attention in recent years, and is undoubtedly more mature than the use of LLM prompting for this task. We have implemented the prompt-based NER techniques recently published in the literature, to the best of our knowledge. It is, of course, possible that new approaches will make it possible to increase performance in the future. However, this is arguably a difficult task for a generative model, as it is highly constrained in its syntax and its evaluation. These results are no indication of performance on other tasks such as classification. 

\subsection{Practical use of language models for low-resource NER}
Overall, our experiments suggest that the performance of language models for clinical named entity recognition is currently sub-optimal. In particular, even MLM-based models, simply fine-tuned on the limited data available, fail to approach the performance of fully supervised models. The three large models trained with the entirety of each training dataset (\emph{skylines} Table~\ref{tab:results}) systematically outperform the best few-shot results, by 5\% to 16\% for the general domain, and 8\% to 48\% for the biomedical domain. However, performance can be judged satisfactory enough for pre-annotation use, to complement or accelerate manual annotation, for example in an online or active learning context.

\section{Conclusion}

This study assessed the performance of two types of large languages models, for few-shot entity recognition in three languages. Our experiments show that few-shot learning performance is significantly lower in the clinical vs. general domain. While masked language models perform better than causal language models (higher F1, lower CO2 emissions), LLM prompting is not yet suited for effective information extraction. 

\section*{Limitations of our study}

\paragraph{Random Noise and Significance}
In MLM experiments, the parameters of the NER tagging layer added on top of the pretrained language model are initialized randomly. Likewise, in LLM prompting experiments, the demonstrations in the prompts are shuffled randomly, and the negative examples in the self verification prompts are selected randomly. These random decisions can introduce noise in our performance measurements. 
Replicating all the experiments would allow us to draw more solid conclusions \cite{reimers-gurevych-EMNLP2017}, but would also come at a considerable cost (29kg of CO2 equivalent, and around 64 hours of computation for each replication). 
The large number of models tested and tasks addressed can however support the main observations of this article. 
For instance, we use Almost Stochastic Order (ASO) \footnote{Given the performance scores of two algorithms A and B, each of which run several times with different settings, ASO computes a test-specific value ($\epsilon_{min}$) that indicates
how far algorithm A is from being significantly better than algorithm B. 
If distance $\epsilon_{min}$ = 0.0, one can claim that A stochastically dominant over B with the predefined significance level. The literature commonly interprets $\epsilon_{min}$ < 0.5 as an indicator of a significant superiority of A over B.}
\cite{dror2019deep} with a confidence level $\alpha = 0.05$ to measure the significance of the superiority of fine-tuned MLMs over prompted LLMs for each dataset separately.
We do not always observe satisfying values of $\epsilon_{min}$ as to whether MLMs perform better than prompted LLMs on general-domain NER (0.426, 0.081 and 0.028 respectively on WikiNER-English, CoNLL2003 and WikiNER-French).
Regarding clinical NER, MLMs perform significantly better than prompted LLMs : MLMs are stochastically dominant over prompted LLMs ($\epsilon_{min}$=0) for all clinical datasets.

\paragraph{Data contamination} The size of the training corpora used for creating LLMs makes it increasingly difficult to control for data contamination, i.e. the presence of test corpora. Moreover, \cite{balloccu2024leak} describe the problem of "indirect contamination", encountered when models are iteratively improved by using data coming from users, including test dataset.

The community is calling for efforts towards better documentation of training datasets \cite{bender-friedman-2018-data}. While some datasets are by construction incompatible with some models (e.g., there is no Spanish training corpus in GPT-J or Llama-2) we are unable to affirm full exclusion of all datasets from all models studied.

\paragraph{Fine-tuning generative models}
Our study focuses on low-data scenarios prevalent in real-world biomedical applications, where computational resources, notably GPU availability, are constrained. Additionally, the use of models via API calls is challenging due to privacy and security concerns.
Consequently, we selected methods requiring limited computing power : either fine-tuning "small", BERT-like models, or prompting larger, frozen, models. This pragmatic choice stems from the need to balance performance with computational efficiency, ensuring that the proposed methods remain feasible for implementation in resource-limited settings.

However, more computationally expensive solutions may exist and may be effectively deployed in other environments. Typically, one could use the limited available data to partially fine-tune a generative LLM for NER \cite{liao2023parameter}.

\paragraph{Error analysis}

To facilitate comparison between models, we evaluated models over one global information extraction performance score: the micro-F1 measure.
Overall, precision and recall scores (not shown) are balanced across entity types. In clinical datasets, some entity types, such as chemicals and anatomy, yield higher performance than others such as devices and procedures. This disparity cannot be explained by the distribution in the training corpus, since prompts provide the same number of examples across entity types. We hypothesize that the regularity and prevalence in LLM pretaining corpora of some entities may explain the higher performance. In future work, we plan more in-depth analysis, to better understand the performance of the LLM prompting approach.

\section*{Ethical considerations}
All the datasets analyzed in our experiments are publicly available corpora, used consistently with the relevant data use agreements.

The use of LLMs can incur significant environmental impact. We have measured carbon emissions using GreenAlgorithms. Our experiments support the conclusion that MLMs yield higher information extraction performance with lower carbon emission, compared to LLM prompting for NER.

\section*{Acknowledgements}

This work has received funding from the French "Agence Nationale pour la Recherche" under grant agreement CODEINE ANR-20-CE23-0026-01.
This work was performed using HPC resources from GENCI-IDRIS (Grant 2023-AD011014533).
The authors thank Dr. Juan Manual Coria for his help phrasing prompts in Spanish.


\bibliography{biblio}
\bibliographystyle{acl_natbib}

\clearpage

\appendix

\section{Appendix}
\subsection{Ablation}
\label{appendices-ablation}

To better understand the contribution of each step of our approach, we carried out a series of complementary experiments.

\subsubsection{Sample and sample size} 
\label{ablation-sample}
We tested our approach with different samples and different sample sizes for one MLM : XLM-RoBERTa-large, and one prompted LLM : Mistral-7B. The results are reported in table \ref{tab:sampling}. It can be noted that, whereas the standard deviation with respect to $p$ is rather high, a significant difference can still be consistently observed between the two models across samples of the same size. We also observe that, as the number of annotated instances decreases, the performance of the MLM drops faster than that of the prompted LLM.

\begin{table}[H]
\centering
\scalebox{0.65}{\begin{tabular}{l|ccc|ccc}
\toprule
& \multicolumn{3}{c|}{CoNLL2003} & \multicolumn{3}{c}{n2c2} \\
\midrule
\multicolumn{5}{c}{\textit{100 annotated instances}} \\
\midrule
 & \textit{p=1} & \textit{p=2} & \textit{p=3} & \textit{p=1} & \textit{p=2} & \textit{p=3} \\
\midrule
\midrule
Mistral-7B & 0.646 & 0.626 & 0.714 & 0.291 & 0.178 & 0.215 \\
XLM-R-large & 0.826 & 0.814 & 0.786 & 0.462 & 0.478 & 0.526 \\
\midrule
\midrule
\multicolumn{5}{c}{\textit{50 annotated instances}} \\
\midrule
 & \textit{p=1} & \textit{p=2} & \textit{p=3} & \textit{p=1} & \textit{p=2} & \textit{p=3} \\
\midrule
\midrule
Mistral-7B & 0.615 & 0.648 & 0.637 & 0.278 & 0.176 & 0.106 \\
XLM-R-large & 0.697 & 0.77 & 0.714 & 0.431 & 0.476 & 0.35 \\
\midrule
\midrule
\multicolumn{5}{c}{\textit{25 annotated instances}} \\
\midrule
 & \textit{p=1} & \textit{p=2} & \textit{p=3} & \textit{p=1} & \textit{p=2} & \textit{p=3} \\
\midrule
\midrule
Mistral-7B & 0.509 & 0.599 & 0.52 & 0.152 & 0.252 & 0.116 \\
XLM-R-large & 0.487 & 0.588 & 0.637 & 0.393 & 0.361 & 0.283 \\
\bottomrule
\end{tabular}}\caption{F1 scores obtained over experiments with different training samples and different training sample sizes.}
\label{tab:sampling}
\end{table}

\subsubsection{Listing prompts} \label{ablation-listing}
In this section, we compare the adopted tagging prompts to listing prompts. In listing prompts, demonstrations simply list the tagged mentions. The list separator is optimized (in the same way as the taggers) between a comma and a newline character. Eventually, the introductory sentences asks to list entities.
The results shown in table \ref{tab:listing} further corroborate our choice of only focusing on tagging prompts.

\begin{table*}[htbp]
\centering
\centerline{\scalebox{0.65}{%
\begin{tabular}{l|ccccc|ccccc|cccc}
 & \multicolumn{5}{c|}{English} & \multicolumn{5}{c|}{French} & \multicolumn{4}{c}{Spanish} \\
\cmidrule{2-6} \cmidrule{7-11} \cmidrule{12-15}
 Model & WikiNER & CoNLL2003 & E3C & n2c2 & NCBI & WikiNER & QFP & E3C & EMEA & MEDLINE & WikiNER & CoNLL2002 & E3C & CWL \\
\midrule
\midrule
 \multicolumn{15}{l}{\textit{Listing prompts}} \\
\midrule
Mistral-7B & 0.659 & 0.533 & 0.417 & 0.281 & 0.340 & 0.676 & 0.083 & 0.451 & 0.169 & 0.403 & 0.697 & 0.620 & 0.211 & 0.273 \\
\midrule
\midrule
\multicolumn{15}{l}{\textit{Tagging prompts}} \\
\midrule
Mistral-7B & 0.754 & 0.646 & 0.488 & 0.291 & 0.395 & 0.727 & 0.428 & 0.590 & 0.229 & 0.333 & 0.720 & 0.707 & 0.083 & 0.374 \\
\bottomrule
\end{tabular}}}
\caption{F1 scores obtained with the listing and tagging prompts.}
\label{tab:listing}
\end{table*}

\subsubsection{Number of demonstrations}
\label{ablation-number}

The decision to limit the number of annotated examples presented to generative models in the prompt to a maximum of 10 is dictated by two constraints.

Firstly, models impose a limit on the number of input tokens over which attention is calculated. Most models used have a limit of 2048 tokens, but more permissive models, such as Mistral-7B, allow up to 8096 tokens. This constraint translates into a limit on the number of sentences that can be presented in the prompt, ranging from 40 to 50 for Mistral-7B and from 10 to 15 for less permissive models, depending on the corpora and tokenizers. For instance, consider the task of detecting body parts in the French section of E3C. Using Mistral-7B (and its tokenizer), the practical limit is around 40 examples, resulting in an average prompt of 7779.5 tokens. Using Vicuna-13B (and its tokenizer), the limit is around 11 examples, resulting in an average of 1851.5 tokens in the prompt.

Secondly, the improvement brought by adding more examples does not appear to be significant, as observed in Table \ref{tab:more}, which shows the results obtained with Mistral-7B when the number of annotated examples is tripled. It is noteworthy that the marginal improvements achieved by tripling the number of examples come at a considerable cost, especially given the quadratic complexity with respect to the length of the prompt.

Therefore, we choose to limit the prompt to 5-10 examples, selected based on the chosen criteria (TF-IDF proximity to the reference sentence or the number of entities present). Instead of selecting examples independently, \citet{gupta2023coverage} propose selecting them interdependently to improve the representativeness of the prompt. This approach would be interesting to implement in our system in future work.
\begin{table*}[htbp]
\centering
\centerline{\scalebox{0.65}{%
\begin{tabular}{l|ccccc|ccccc|cccc}
 & \multicolumn{5}{c|}{English} & \multicolumn{5}{c|}{French} & \multicolumn{4}{c}{Spanish} \\
\cmidrule{2-6} \cmidrule{7-11} \cmidrule{12-15}
 Model & WikiNER & CoNLL2003 & E3C & n2c2 & NCBI & WikiNER & QFP & E3C & EMEA & MEDLINE & WikiNER & CoNLL2002 & E3C & CWL \\
\midrule
\midrule
\multicolumn{15}{l}{\textit{5/10 demonstrations}} \\
\midrule
Mistral-7B & 0.754 & 0.646 & 0.488 & 0.291 & 0.395 & 0.727 & 0.428 & 0.590 & 0.229 & 0.333 & 0.720 & 0.707 & 0.083 & 0.374 \\
\midrule
\midrule
\multicolumn{15}{l}{\textit{15/30 demonstrations}} \\
\midrule
Mistral-7B & 0.763 & 0.692 & 0.453 & 0.263 & 0.377 & 0.782 & 0.355 & 0.587 & 0.237 & 0.396 & 0.785 & 0.751 & 0.163 & 0.413 \\
\bottomrule
\end{tabular}}}
\caption{F1 scores obtained with 5/10 demonstations vs. with 15/30 demonstations}
\label{tab:more}
\end{table*}

\subsubsection{Hyperparameter grid search}
\label{ablation-grid}
In order to assess the quality of our adopted search method used to find the best feature combination to incorporate in the prompt, we compare this method to a naïve grid search over these features. We test all 512 combinations of our identified 9 features, for Mistral-7B over ConLL2003. The scores found through LOOCV vary between 0.0 and 0.656 with a mean value of 0.387 and a median of 0.46. The best-preforming combination is : \textit{Additional sentences}, \textit{Self-verification}, \textit{Introductory sentence for the test instance} and \textit{Require a long answer for the self-verification}, which is exactly the same combination we found initially through a greedy, tree search, that is around 20 times faster and less consuming.

\subsection{Evaluated models}
\label{appendices-models}
Table \ref{tab:LM_features} specifies relevant information about the tested models. In the training corpus column, we point out the data contaminations we know of. For instance, MedBERT's documentation explicitly mentions N2C2 as part of its training data. This might lead to artificially high evaluation metrics, as the model is not generalizing to unseen data, but instead leveraging pre-learned information, which compromises the validity of the results in real-world applications. 
\begin{table*}[htbp]
\centerline{\scalebox{0.75}{
\begin{tabular}{clp{7cm}rp{3cm}p{10cm}}
\toprule
& \# & Model & \makecell[r]{Number of\\ parameters} & Training data size & \makecell[l]{Training corpus} \\
\midrule[2pt]
\multirow{13}{*}{\rotatebox[origin=c]{90}{Causal}} & 1 & Llama-2-70B\textsuperscript{\texttt{[en]}} \cite{touvron2023llama} & 70B & 2 trillion tokens & A mix of publicly available online data, mainly in English \\
\cmidrule{2-6}
 & 2 & Llama-3-8B-Instruct\textsuperscript{\texttt{[en]}} & 8B & over 15 trillion tokens & "A new mix" of publicly available online data, mainly in English \\
\cmidrule{2-6}
 & 3 & Mistral-7B\textsuperscript{\texttt{[?]}} \cite{jiang2023mistral} & 7B & Undisclosed & Undisclosed \\
\cmidrule{2-6}
 & 4 & Phi-3-medium-instruct\textsuperscript{\texttt{[en]}} \cite{abdin2024phi} & 14B & 4.8 trillion tokens & A combination of publicly available corpora, synthetic data and chat format supervised data, mainly in English \\
\cmidrule{2-6}
 & 5 & BLOOM-7B1\textsuperscript{\texttt{[en]}}\textsuperscript{\texttt{[fr]}}\textsuperscript{\texttt{[es]}} \cite{workshop2022bloom} & 7B & 1.6 TB & ROOTS \cite{laurenccon2022bigscience}, a mix of datasets and pseudo-crawled data 59 languages \\
\cmidrule{2-6}
 & 6 & Falcon-40B\textsuperscript{\texttt{[en]}}\textsuperscript{\texttt{[fr]}}\textsuperscript{\texttt{[es]}} & 40B & 1 trillion tokens & RefinedWeb \cite{penedo2023refinedweb}, a dataset of filtered and deduplicated web data \\
\cmidrule{2-6}
 & 7 & GPT-J-6B\textsuperscript{\texttt{[en]}} \cite{wang2021gptj} & 6B & 825 GiB & The Pile \cite{gao2020pile}, a mixture of public datasets and web data in English \\
\cmidrule{2-6}
 & 8 & OPT-66B\textsuperscript{\texttt{[en]}} \cite{zhang2022opt} & 66B & 180 billion tokens & Crawled data from the web, mainly in English \\
\cmidrule{2-6}
 & 9 & Vicuna-13B\textsuperscript{\texttt{[en]}}* \cite{zheng2023judging} & 13B & 125K conversations & Llama 2, fine-tuned on conversations collected from ShareGPT.com, mainly in English \\
\cmidrule{2-6}
 & 10 & Vicuna-7B\textsuperscript{\texttt{[en]}}* \cite{zheng2023judging} & 7B & 125K conversations & Llama 2, fine-tuned on conversations collected from ShareGPT.com, mainly in English \\
\cmidrule{2-6}
 & 11 & BioMistral-7B\textsuperscript{\texttt{[en]}}* \cite{labrak2024biomistral} & 7B & 3 billion tokens & Mistral, fine-tuned on the PMC Open Access Subset \\
\cmidrule{2-6}
 & 12 & Medalpaca-7B\textsuperscript{\texttt{[en]}}* \cite{han2023medalpaca} & 7B & 400K Q.A. pairs & Llama 2, fine-tuned on semi-generated medical question-answer pairs in English \\
\cmidrule{2-6}
 & 13 & Vigogne-13B\textsuperscript{\texttt{[fr]}}\textsuperscript{\texttt{[en]}}* & 13B & 52K instructions & Llama 2, fine-tuned on English instructions automatically translated to French \\
\midrule[2pt]
\multirow{16}{*}{\rotatebox[origin=c]{90}{Masked}} & 14 & mBERT\textsuperscript{\texttt{[en]}}\textsuperscript{\texttt{[fr]}}\textsuperscript{\texttt{[es]}} \cite{devlin2019bert} & 110M & Undisclosed & A corpus featuring 104 languages built from undisclosed sources \\
\cmidrule{2-6}
 & 15 & XLM-R-large\textsuperscript{\texttt{[en]}}\textsuperscript{\texttt{[fr]}}\textsuperscript{\texttt{[es]}} \cite{conneau2020unsupervised} & 355M & 2.5 TB & Filtered CommonCrawl data containing 100 languages \\
\cmidrule{2-6}
 & 16 & BERT-large\textsuperscript{\texttt{[en]}} \cite{devlin2019bert} & 345M & 3,3 billion words & BookCorpus \cite{zhu2015aligning}, a dataset consisting of unpublished books and English Wikipedia. \\
\cmidrule{2-6}
 & 17 & RoBERTa-large\textsuperscript{\texttt{[en]}} \cite{liu2019roberta} & 355M & 160 GiB & BooksCorpus \cite{zhu2015aligning}, English Wikipedia, and crawled web data \\
\cmidrule{2-6}
 & 18 & Bio\_ClinicalBERT\textsuperscript{\texttt{[en]}} \cite{alsentzer2019publicly} & 110M & 2 million clinical notes & MIMIC-III \cite{johnson2016mimic}, a database containing electronic health records from hospitalized ICU patients \\
\cmidrule{2-6}
 & 19 & ClinicalBERT\textsuperscript{\texttt{[en]}} \cite{wang2023optimized} & 110M & 1.2 billion words & A large multi-center dataset with a corpus built from undisclosed sources \\
\cmidrule{2-6}
 & 20 & MedBERT\textsuperscript{\texttt{[en]}} \cite{charangan2022medbert} & 110M & 57 million words & Community datasets (including N2C2 \cite{luo2020n2c2}) and Crawled medical-related articles from Wikipedia \\
\cmidrule{2-6}
 & 21 & CamemBERT-large\textsuperscript{\texttt{[fr]}} \cite{martin2019camembert} & 335M & 64 billion tokens & OSCAR \cite{suarez2020monolingual}, a corpus of web data in French \\
\cmidrule{2-6}
 & 22 & FlauBERT-large\textsuperscript{\texttt{[fr]}} \cite{le2019flaubert} & 335M & 13 billion tokens & A mix of French Wikipedia, French books, and French web data \\
\cmidrule{2-6}
 & 23 & DrBERT-4GB\textsuperscript{\texttt{[fr]}} \cite{labrak2023drbert} & 110M & 1 billion words & A mix of publicly available biomedical corpora in French (including QuaeroFrenchMed \cite{neveol2014quaero}). \\
\cmidrule{2-6}
 & 24 & CamemBERT-bio\textsuperscript{\texttt{[fr]}} \cite{touchent2023camembertbio} & 110M & 413 million words & A mix of publicly available biomedical corpora in French (including E3C \cite{magnini2021e3c}). \\
\cmidrule{2-6}
 & 25 & BETO\textsuperscript{\texttt{[es]}} \cite{canete2020beto} & 110M & 3 billion words & Spanish Wikipedia and Spanish data from OPUS \cite{tiedemann2012parallel} \\
\cmidrule{2-6}
 & 26 & PatanaBERT\textsuperscript{\texttt{[es]}} & 110M & Undisclosed & Spanish \\
\cmidrule{2-6}
 & 27 & TulioBERT\textsuperscript{\texttt{[es]}} & 110M & Undisclosed & Spanish \\
\cmidrule{2-6}
 & 28 & BSC-BioEHR\textsuperscript{\texttt{[es]}} \cite{carrino2022pretrained} & 110M & 1.1 billion tokens & A mixture of biomedical community datasets including EHR documents and crawled data in Spanish \\
\cmidrule{2-6}
 & 29 & BSC-Bio\textsuperscript{\texttt{[es]}} \cite{carrino2022pretrained} & 110M & 963 million tokens & A mixture of biomedical community datasets and crawled data in Spanish \\
\bottomrule
\end{tabular}}}
\caption{Characterization of the language models used in our experiments in terms of parameters and training corpus. Models marked with \textsuperscript{\texttt{[en]}} (respectively \textsuperscript{\texttt{[fr]}}, \textsuperscript{\texttt{[es]}}) are heavily trained on English (respectively French, Spanish). CLMs marked with * are fine-tuned versions of other CLMs.}
\label{tab:LM_features}
\end{table*}

\subsection{NER labels descriptions}
Tables \ref{tab:ner_tags_en} to \ref{tab:ner_tags_es2} definitions of all the labels present in the studied datasets. These definitions were drawn from available annotation schemes and further curated by the authors and native speakers. These definitions were included in the prompt when the "Include label definitions in the prompt" feature was activated.
\label{appendices-descriptions}
\begin{table*}
\centering
\centerline{\scalebox{0.8}{\begin{tabular}{p{3cm}p{4cm}p{12cm}}
\toprule
Tag & Tag name (in singular) & Description \\
\midrule[2pt]
PER & person names (a person's name)  & These are names of persons such as real people or fictional characters.  \\
\midrule
FAC & facilities (a facility)  & These are names of man-made structures such as infrastructure, buildings and monuments.  \\
\midrule
LOC & locations (a location)  & These are names of geographical locations such as landmarks, cities, countries and regions.  \\
\midrule
ORG & organizations (an organization)  & These are names of organizations such as companies, agencies and political parties.  \\
\midrule
FUNC & functions and jobs (a function or a job)  & These are words that refer to a profession or a job.  \\
\midrule
ACTI & activities and behaviors (an activity or behavior)  & These are words that refer to human activities, behaviors or events as well as governmental or regulatory activities.  \\
\midrule
ANAT & anatomy (an anatomy)  & These are words that refer to the structure of the human body, its organs and their position, such as body parts or organs, systems, tissues, cells, body substances and embryonic structures.  \\
\midrule
CHEM & chemicals and drugs (a chemical or a drug)  & These are words that refer to a substance or composition that has a chemical characteristic, especially a curative or preventive property with regard to human or animal diseases, such as drugs, antibiotics, proteins, hormones, enzymes and hazardous or poisonous substances.  \\
\midrule
CONC & concepts and ideas (a concept or an idea)  & These are words that refer to a concept or an idea, such as a classification, an intellectual product, a language, a law or a regulation.  \\
\midrule
DEVI & medical devices (a device)  & These are words that refer to a medical device used to administer care or perform medical research.  \\
\midrule
DISO & disorders (a disorder)  & These are words that refer to an alteration of morphology, function or health of a living organism, animal or plant, such as congenital abnormalities, dysfunction, injuries, signs or symptoms or observations.  \\
\midrule
GENE & genes and molecular sequences (a gene or a molecular sequence)  & These are words that refer to a gene, a genome or a molecular sequence.  \\
\midrule
GEOG & geographical areas (a geographical area)  & These are words that refer to a country, a region or a city.  \\
\midrule
LIVB & living beings (a living being)  & These are words that refer to a living being or a group of living beings, such as a person or a group of persons, a plant or a category of plants, an animal or a category of animals.  \\
\midrule
OBJC & objects (an object)  & These are words that refer to anything animate or inanimate that affects the senses, such as physical manufactured objects.  \\
\midrule
OCCU & occupations (an occupation)  & These are words that refer to a professional occupation or discipline.  \\
\midrule
ORGA & organizations (an organization)  & These are words that refer to an organization such as healthcare related organizations.  \\
\midrule
PHEN & phenomema (a phenomemon)  & These are words that refer to a phenomenon that occurs naturally or as a result of an activity, such as a biologic function.  \\
\midrule
PHYS & physiology (a physiology)  & These are words that refer to any element that contributes to the mechanical, physical and biochemical functioning or organization of living organisms and their components.  \\
\midrule
PROC & procedures (a procedure)  & These are words that refer to an activity or a procedure that contributes to the diagnosis or treatment of patients, the information of patients, the training of medical personnel or biomedical research.  \\
\midrule
EVENT & events (an event)  & These are words that refer to actions, states, and circumstances that are relevant to the clinical history of a patient such as pathologies and symptoms, or more generally words like "enters", "reports" or "continue".  \\
\bottomrule
\end{tabular}}}
\caption{Description of the NER tags used in our experiments for English.}
\label{tab:ner_tags_en}
\end{table*}

\begin{table*}
\centering
\centerline{\scalebox{0.8}{\begin{tabular}{p{3cm}p{4cm}p{12cm}}
\toprule
Tag & Tag name (in singular) & Description \\
\midrule[2pt]
TIMEX3 & time expressions (a time expression)  & These are time expressions such as dates, times, durations, frequencies, or intervals.  \\
\midrule
RML & results and measurements (a result or a measurement)  & These are test results, results of laboratory analyses, formulaic measurements, and measure values.  \\
\midrule
ACTOR & actors (an actor)  & These are words that refer patients, healthcare professionals, or other actors relevant to the clinical history of a patient.  \\
\midrule
Abbreviation & abbreviations (an abbreviation)  & These are words that refer to abbreviations.  \\
\midrule
Body\_Part & body parts (a body part)  & These are words that refer to organs and anatomical parts of persons.  \\
\midrule
Clinical\_Finding & clinical findings (a clinical finding)  & These are words that refer to observations, judgments or evaluations made about patients.  \\
\midrule
Diagnostic\_Procedure & diagnostic procedures (a diagnostic procedure)  & These are words that refer to tests that allow determining the condition of the individual.  \\
\midrule
Disease & diseases (a disease)  & These are words that describe an alteration of the physiological state in one or several parts of the body, due to generally known causes, manifested by characteristic symptoms and signs, and whose evolution is more or less predictable.  \\
\midrule
Family\_Member & family members (a family member)  & These are words that refer to family members.  \\
\midrule
Laboratory\_or \\\_Test\_Result & laboratory or test results (a laboratory or test result)  & These are words that refer to any measurement or evaluation obtained from a diagnostic support examination.  \\
\midrule
Laboratory\_Procedure & laboratory procedures (a laboratory procedure)  & These are words that refer to tests that are performed on various patient samples that allow diagnosing diseases by detecting biomarkers and other parameters. Blood, urine, and other fluids and tissues that use biochemical, microbiological and/or cytological methods are considered.  \\
\midrule
Medication & medications (a medication)  & These are words that refer to medications or drugs used in the treatment and/or prevention of diseases, including brand names and generics, as well as names for groups of medications.  \\
\midrule
Procedure & procedures (a procedure)  & These are words that refer to activities derived from the care and care of patients.  \\
\midrule
Sign\_or\_Symptom & signs or symptoms (a sign or symptom)  & These are words that refer to manifestations of a disease, determined by medical examination or perceived and expressed by the patient.  \\
\midrule
Therapeutic\_Procedure & therapeutic procedures (a therapeutic procedure)  & These are words that refer to activities or treatments that are used to prevent, repair, eliminate or cure the individual's disease.  \\
\midrule
CompositeMention & composite mentions of diseases (a composite mention of diseases)  & These are words that refer to mentions of multiple diseases, such as "colorectal, endometrial, and ovarian cancers".  \\
\midrule
DiseaseClass & disease classes (a disease class)  & These are words that refer to classes of diseases, such as "an autosomal recessive disease".  \\
\midrule
Modifier & modifiers (a modifier of diseases)  & These are words that refer to modifiers of diseases, such as "primary" or "C7-deficient".  \\
\midrule
SpecificDisease & diseases (a disease)  & These are words that refer to specific diseases, such as "diastrophic dysplasia".  \\
\bottomrule
\end{tabular}}}
\caption{Description of the NER tags used in our experiments for English, continued.}
\label{tab:ner_tags_en2}
\end{table*}

\begin{table*}
\centering
\centerline{\scalebox{0.8}{\begin{tabular}{p{3cm}p{4cm}p{12cm}}
\toprule
Tag & Tag name (in singular) & Description \\
\midrule[2pt]
PER & de noms de personnes (un nom de personne)  & Il s'agit des noms de personnes, qu'elles soient réelles ou fictives.  \\
\midrule
FAC & de productions humaines (une production humaine)  & Il s'agit des noms de structures faites par les humains comme des infrastructures, des bâtiments ou des monuments.  \\
\midrule
LOC & de lieux (un lieu)  & Il s'agit des noms de lieux comme des endroits, villes, pays ou régions.  \\
\midrule
ORG & d'organisations (une organisation)  & Il s'agit des noms d'organisations comme des entreprises, des agences ou des partis politiques.  \\
\midrule
FUNC & de fonctions et métiers (une fonction ou un métier)  & Il s'agit de mots qui se rapportent à une activité professionnelle.  \\
\midrule
ANAT & d'anatomie (une partie du corps)  & Il s'agit d'une entité se rapportant à la structure du corps humain, ses organes et leur position. Il s’agit principalement des parties du corpus ou organes, des appareils, des tissus, des cellules, des substances corporelles et des organismes embryonaires.  \\
\midrule
CHEM & de médicaments et substances chimiques (un médicament ou une substance chimique)  & Il s'agit d'une substance ou composition présentant des propriétés chimiques caractéristiques, en particulier des propriétés curatives ou préventives à l’égard des maladies humaines ou animales. Il s’agit principalement des médicaments disponibles en pharmacie, des antibiotiques, des proteines, des hormones, des substances dangereuses, des enzymes.  \\
\midrule
DEVI & de matériel (un matériel)  & Il s'agit d'un matériel utilisé pour administrer des soins ou effectuer des recherches médicales.  \\
\midrule
DISO & de problèmes médicaux (un problème médical)  & Il s'agit d'une altération de la morphologie, des fonctions, ou de la santé d’un organisme vivant, animal ou végétal. Il peut s’agir de malformations, de maladies, de blessure, de signe ou symptome ou d’une observation.  \\
\midrule
GEOG & de zones géographiques (une zone géographique)  & Il s'agit d'un pays, une région, ou une ville.  \\
\midrule
LIVB & d'êtres vivants (un être vivant)  & Il s'agit d'un être vivant ou groupe d’êtres vivants. Il peut s’agir d’une personne ou d’un groupe de personnes, d’une plante ou d’une catégorie de végétaux, d’un animal ou d’une catégorie d’animaux.  \\
\midrule
OBJC & d'objets (un objet)  & Il s'agit de tout ce qui, animé ou inanimé, affecte les sens. Ici, il s’agit principalement d’objets physiques manufacturés.  \\
\midrule
PHEN & de phénomènes (un phénomène)  & Il s'agit d'un phénomène qui se produit naturellement ou à la suite d’une activité. Il s’agit principalement de fonctions biologiques.  \\
\midrule
PHYS & de physiologie (une physiologie)  & Il s'agit de tout élément contribuant au fonctionnement ou à l’organisation mécanique, physique et biochimique des organismes vivants et de leurs composants.  \\
\midrule
PROC & de procédures (une procédure)  & Il s'agit d'une activité ou procédure contribuant au diagnostic ou au traitement des patients, à l’information des patients, la formation du personnel médical ou à la recherche biomédicale.  \\
\midrule
EVENT & d'événements (un événement)  & Il s'agit d'une action, d’un état ou d’une circonstance qui est pertinent pour l’histoire clinique d’un patient. Il peut s’agir de pathologies et symptômes, ou plus généralement de mots comme "entre", "rapporte" ou "continue".  \\
\midrule
TIMEX3 & d'expressions temporelles (une expression temporelle)  & Il s'agit d’expressions temporelles comme des dates, heures, durées, fréquences, ou intervalles.  \\
\midrule
RML & de résultats et mesures (un résultat ou une mesure)  & Il s'agit de résultats d’analyses de laboratoire, de mesures formelles, et de valeurs de mesure.  \\
\midrule
ACTOR & d'acteurs (un acteur)  & Il s'agit de patients, de professionnels de santé, ou d’autres acteurs pertinents pour l’histoire clinique d’un patient.  \\
\bottomrule
\end{tabular}}}
\caption{Description of the NER tags used in our experiments for French.}
\label{tab:ner_tags_fr}
\end{table*}
\begin{table*}
\centering
\centerline{\scalebox{0.8}{\begin{tabular}{p{3cm}p{4cm}p{12cm}}
\toprule
Tag & Tag name (in singular) & Description \\
\midrule[2pt]
PER & nombres de personas (un nombre de persona)  & Estos son nombres de personas, ya sean reales o personajes ficticios.  \\
\midrule
FAC & instalaciones (una instalación)  & Estos son nombres de estructuras hechas por el hombre como infraestructura, edificios y monumentos.  \\
\midrule
LOC & lugares (un lugar)  & Estos son nombres de ubicaciones geográficas como hitos, ciudades, países y regiones.  \\
\midrule
ORG & organizaciones (una organización)  & Estos son nombres de organizaciones como empresas, agencias y partidos políticos.  \\
\midrule
ACTI & actividades y comportamientos (una actividad o comportamiento)  & Estas son palabras que se refieren a actividades humanas, comportamientos o eventos, así como actividades gubernamentales o regulatorias.  \\
\midrule
ANAT & anatomía (una anatomía)  & Estas son palabras que se refieren a la estructura del cuerpo humano, sus órganos y su posición, como partes del cuerpo u órganos, sistemas, tejidos, células, sustancias corporales y estructuras embrionarias.  \\
\midrule
CHEM & productos químicos y medicamentos (un producto químico o un medicamento)  & Estas son palabras que se refieren a una sustancia o composición que tiene una característica química, especialmente una propiedad curativa o preventiva con respecto a las enfermedades humanas o animales, como medicamentos, antibióticos, proteínas, hormonas, enzimas y sustancias peligrosas o venenosas.  \\
\midrule
CONC & conceptos e ideas (un concepto o una idea)  & Estas son palabras que se refieren a un concepto o una idea, como una clasificación, un producto intelectual, un idioma, una ley o un reglamento.  \\
\midrule
DEVI & dispositivos médicos (un dispositivo)  & Estas son palabras que se refieren a un dispositivo médico utilizado para administrar atención o realizar investigaciones médicas.  \\
\midrule
DISO & trastornos (un trastorno)  & Estas son palabras que se refieren a una alteración de la morfología, la función o la salud de un organismo vivo, animal o vegetal, como anomalías congénitas, disfunción, lesiones, signos o síntomas u observaciones.  \\
\midrule
GENE & genes y secuencias moleculares (un gen o una secuencia molecular)  & Estas son palabras que se refieren a un gen, un genoma o una secuencia molecular.  \\
\midrule
GEOG & áreas geográficas (un área geográfica)  & Estas son palabras que se refieren a un país, una región o una ciudad.  \\
\midrule
LIVB & seres vivos (un ser vivo)  & Estas son palabras que se refieren a un ser vivo o un grupo de seres vivos, como una persona o un grupo de personas, una planta o una categoría de plantas, un animal o una categoría de animales.  \\
\midrule
OBJC & objetos (un objeto)  & Estas son palabras que se refieren a cualquier cosa animada o inanimada que afecte los sentidos, como objetos físicos fabricados.  \\
\midrule
OCCU & ocupaciones (una ocupación)  & Estas son palabras que se refieren a una ocupación o disciplina profesional.  \\
\midrule
ORGA & organizaciones (una organización)  & Estas son palabras que se refieren a una organización, por ejemplo organizaciones relacionadas con la salud.  \\
\midrule
PHEN & fenómenos (un fenómeno)  & Estas son palabras que se refieren a un fenómeno que ocurre naturalmente o como resultado de una actividad, por ejemplo una función biológica.  \\
\bottomrule
\end{tabular}}}
\caption{Description of the NER tags used in our experiments for Spanish.}
\label{tab:ner_tags_es}
\end{table*}

\begin{table*}
\centering
\centerline{\scalebox{0.8}{\begin{tabular}{p{3cm}p{4cm}p{12cm}}
\toprule
Tag & Tag name (in singular) & Description \\
\midrule[2pt]
PHYS & fisiología (una fisiología)  & Estas son palabras que se refieren a cualquier elemento que contribuya al funcionamiento mecánico, físico y bioquímico o la organización de los organismos vivos y sus componentes.  \\
\midrule
PROC & procedimientos (un procedimiento)  & Estas son palabras que se refieren a una actividad o un procedimiento que contribuye al diagnóstico o tratamiento de pacientes, la información de pacientes, la capacitación del personal médico o la investigación biomédica.  \\
\midrule
EVENT & eventos (un evento)  & Estas son palabras que se refieren a acciones, estados y circunstancias que son relevantes para la historia clínica de un paciente, como patologías y síntomas, o más generalmente palabras como "entra", "reporta" o "continúa".  \\
\midrule
TIMEX3 & expresiones de tiempo (una expresión de tiempo)  & Estas son expresiones de tiempo como fechas, horas, duraciones, frecuencias o intervalos.  \\
\midrule
RML & resultados y mediciones (un resultado o una medida)  & Estos son resultados de análisis de laboratorio, mediciones formales y valores de medición.  \\
\midrule
ACTOR & actores (un actor)  & Estas son palabras que se refieren a pacientes, profesionales de la salud u otros actores relevantes para la historia clínica de un paciente.  \\
\midrule
Abbreviation & abreviaciones (una abreviación)  & Estas son los casos de siglas y acrónimos.  \\
\midrule
Body\_Part & partes del cuerpo (una parte del cuerpo)  & Estas son palabras que se refieren a òrganos y partes anatómicas de personas.  \\
\midrule
Clinical\_Finding & hallazgos clínicos (un hallazgo clínico)  & Estas son palabras que se refieren a observaciones, juicios o evaluaciones que se hacen sobre los pacientes.  \\
\midrule
Diagnostic\_Procedure & procedimientos diagnósticos (un procedimiento diagnóstico)  & Estas son palabras que se refieren a exámenes que permiten determinar la condición del individuo.  \\
\midrule
Disease & enfermedades (una enfermedad)  & Estas son palabras que describen una alteración del estado fisiológico en una o varias partes del cuerpo, por causas en general conocidas, manifestada por síntomas y signos característicos, y cuya evolución es más o menos previsible.  \\
\midrule
Family\_Member & miembros de la familia (un miembro de la familia)  & Estas son palabras que se refieren a miembros de la familia.  \\
\midrule
Laboratory\_or\\
\_Test\_Result & resultados de exámenes de laboratorio u otras pruebas (un resultado de un examen de laboratorio u otra prueba)  & Estas son palabras que se refieren a cualquier medición o evaluación obtenida a partir de un exámen de apoyo diagnóstico.  \\
\midrule
Laboratory\_Procedure & procedimientos de laboratorio (un procedimiento de laboratorio)  & Estas son palabras que se refieren a exámenes que se realizan en diversas muestras de pacientes que permiten diagnosticar enfermedades mediante la detección de biomarcadores y otros parámetros. Se consideran los análisis de sangre, orina, y otros fluidos y tejidos que emplean métodos bioquímicos, microbiológicos y/o citológicos.  \\
\midrule
Medication & medicamentos o drogas (un medicamento o una droga)  & Estas son palabras que se refieren a medicamentos o drogas empleados en el tratamiento y/o prevención de enfermedades, incluyendo marcas comerciales y genéricos, así como también nombres para grupos de medicamentos.  \\
\midrule
Procedure & procedimientos (un procedimiento)  & Estas son palabras que se refieren a actividades derivadas de la atención y el cuidado de los pacientes.  \\
\midrule
Sign\_or\_Symptom & signos o síntomas (un signo o un síntoma)  & Estas son palabras que se refieren a manifestaciones de una enfermedad, determinadas mediante la exploración médica o percibidas y expresadas por el paciente.  \\
\midrule
Therapeutic\_Procedure & procedimientos terapéuticos (un procedimiento terapéutico)  & Estas son palabras que se refieren a actividades o tratamientos que es empleado para prevenir, reparar, eliminar o curar la enfrmedad del individuo.  \\
\bottomrule
\end{tabular}}}
\caption{Description of the NER tags used in our experiments for Spanish, continued.}
\label{tab:ner_tags_es2}
\end{table*}

\subsection{Carbon footprint}
\label{appendices-carbon}
Tables \ref{tab:impact_validation} and \ref{tab:impact_test} detail the carbon emission estimations for all of our experiments. These estimations were computed with GreenAlgorithms v2.2 \cite{lannelongue2021green} \footnote{http://calculator.green-algorithms.org/}, based on factors such as runtime, computing hardware and location where electricity used by our computer facility was produced.
In total, the experiments described in this paper are estimated to have generated around 31kg of CO2 equivalent (29kg for the main experiments, and 2kg for ablation).
\begin{table*}
\centering
\centerline{\scalebox{0.65}{\begin{tabular}{lll|ccccc|ccccc|cccc}
 & & & \multicolumn{5}{c|}{English} & \multicolumn{5}{c|}{French} & \multicolumn{4}{c}{Spanish} \\
\cmidrule{3-7} \cmidrule{8-12} \cmidrule{13-17}
& \# & Model & WikiNER & CoNLL2003 & E3C & n2c2 & NCBI & WikiNER & QFP & E3C & EMEA & MEDLINE & WikiNER & CoNLL2002 & E3C & CWL \\
\midrule
\midrule
 \multicolumn{17}{l}{\textit{Few-shot approaches}} \\
\midrule
\multirow{12}{*}{\rotatebox[origin=c]{90}{Causal}} 
 & 1 & LLAMA-2-70B & 46 & 44 & 126 & 233 & 54 & 85 & 131 & 129 & 273 & 284 & 41 & 76 & 114 & 344 \\
 & 2 & LLAMA-3-8B-Instruct & 3 & 5 & 12 & 19 & 12 & 5 & 7 & 7 & 11 & 23 & 4 & 7 & 12 & 36 \\
 & 3 & Mistral-7B & 4 & 6 & 12 & 24 & 8 & 5 & 8 & 14 & 13 & 25 & 7 & 5 & 11 & 27 \\ 
 & 4 & Phi-3-medium-instruct & 5 & 8 & 14 & 24 & 6 & 12 & 9 & 14 & 17 & 25 & 9 & 15 & 19 & 28 \\ 
 & 5 & BLOOM-7B1 & 4 & 6 & 10 & 26 & 9 & 8 & 13 & 9 & 26 & 20 & 4 & 8 & 8 & 18 \\
 & 6 & Falcon-40B & 49 & 45 & 56 & 176 & 45 & 31 & 58 & 75 & 162 & 129 & 33 & 25 & 82 & 99 \\
 & 7 & GPT-J-6B &7 & 6 & 8 & 23 & 7 & 5 & 8 & 13 & 21 & 17 & 6 & 6 & 13 & 28 \\
 & 8 & OPT-66B & 73 & 50 & 120 & 253 & 96 & 38 & 64 & 138 & 273 & 240 & 57 & 52 & 106 & 247 \\
 & 9 & Vicuna-13B & 10 & 11 & 20 & 52 & 11 & 11 & 12 & 18 & 33 & 40 & 10 & 11 & 22 & 51 \\
 & 10 & Vicuna-7B & 6 & 8 & 14 & 17 & 6 & 5 & 10 & 10 & 24 & 14 & 8 & 6 & 13 & 27 \\
 & 11 & BioMistral-7B & 7 & 8 & 11 & 17 & 5 & 7 & 12 & 10 & 28 & 14 & 8 & 8 & 11 & 23 \\
 & 12 & Medalpaca-7B & 8 & 4 & 17 & 24 & 10 & 7 & 14 & 11 & 19 & 13 & 7 & 8 & 15 & 26 \\
 & 13 & Vigogne-13B & 14 & 14 & 29 & 37 & 11 & 13 & 20 & 26 & 36 & 39 & 11 & 14 & 32 & 44 \\
\midrule
\multirow{16}{*}{\rotatebox[origin=c]{90}{Masked}} & 11 & mBERT & 2 & 1 & 2 & 2 & 2 & 2 & 2 & 2 & 1 & 1 & 1 & 2 & 1 & 2 \\
 & 12 & XLM-R-large & 2 & 2 & 2 & 1 & 2 & 2 & 2 & 2 & 2 & 2 & 1 & 1 & 1 & 2 \\
 & 13 & BERT-large & 2 & 1 & 2 & 2 & 2 & - & - & - & - & - & - & - & - & - \\
 & 14 & RoBERTa-large & 1 & 2 & 2 & 2 & 2 & - & - & - & - & - & - & - & - & - \\
 & 15 & Bio\_ClinicalBERT & 2 & 2 & 1 & 2 & 1 & - & - & - & - & - & - & - & - & - \\
 & 16 & ClinicalBERT & 1 & 1 & 2 & 2 & 1& - & - & - & - & - & - & - & - & - \\
 & 17 & MedBERT & 2 & 2 & 1 & 1 & 1 & - & - & - & - & - & - & - & - & - \\
 & 18 & CamemBERT-large & - & - & - & - & - & 1 & 1 & 1 & 2 & 2 & - & - & - & - \\
 & 19 & FlauBERT-large &- & - & - & - & - & 2 & 2 & 2 & 2 & 2 & - & - & - & - \\
 & 20 & DrBERT-4GB & - & - & - & - & -  & 2 & 2 & 2 & 2 & 2 & - & - & - & - \\
 & 21 & CamemBERT-bio & - & - & - & - & - & 1 & 2 & 2 & 2 & 2 & - & - & - & - \\
 & 23 & BETO & - & - & - & - & - & - & - & - & - & - & 2 & 1 & 1 & 1 \\
 & 23 & PatanaBERT & - & - & - & - & - & - & - & - & - & - & 2 & 2 & 2 & 2 \\
 & 24 & TulioBERT &- & - & - & - & - & - & - & - & - & -  & 1 & 2 & 2 & 1 \\
 & 25 & BSC-BioEHR & - & - & - & - & - & - & - & - & - & -  & 2 & 2 & 2 & 2 \\
 & 26 & BSC-Bio & - & - & - & - & - & - & - & - & - & - & 2 & 2 & 2 & 2 \\
\midrule
\midrule
\multicolumn{17}{l}{\textit{Masked fully-supervised (skyline)}} \\
\midrule
 & & RoBERTa-large & 647 & 68 & 5 & 12 & 24 & - & - & - & - & - & - & - & - & - \\
 & & CamemBERT-large & - & - & - & - & - & 595 & 15 & 4 & 5 & 8 & - & - & - & - \\
 & & BETO & - & - & - & - & - & - & - & - & - & - & 579 & 41 & 3 & 21 \\

\bottomrule
\end{tabular}}}
\caption{This table presents the carbon emissions (in g) of the optimization on the validation set of each model over each dataset. For CLMs, this correponds to the tree search over the prompt features through cross-validation. For MLMs, this corresponds to the supervised fine-tuning and training of the model.}
\label{tab:impact_validation}
\end{table*}

\begin{table*}
\centering
\centerline{\scalebox{0.65}{\begin{tabular}{lll|ccccc|ccccc|cccc}
 & & & \multicolumn{5}{c|}{English} & \multicolumn{5}{c|}{French} & \multicolumn{4}{c}{Spanish} \\
\cmidrule{3-7} \cmidrule{8-12} \cmidrule{13-17}
& \# & Model & WikiNER & CoNLL2003 & E3C & n2c2 & NCBI & WikiNER & QFP & E3C & EMEA & MEDLINE & WikiNER & CoNLL2002 & E3C & CWL \\
\midrule
\midrule
 \multicolumn{17}{l}{\textit{Few-shot approaches}} \\
\midrule
\multirow{12}{*}{\rotatebox[origin=c]{90}{Causal}}
 & 1 & LLAMA-2-70B & 812 & 147 & 36 & 196 & 33 & 508 & 11 & 13 & 92 & 47 & 514 & 201 & 11 & 198 \\
 & 2 & LLAMA-3-8B-Instruct & 240 & 39 & 9 & 63 & 22 & 158 & 3 & 4 & 29 & 21 & 279 & 53 & 2 & 34\\
 & 3 & Mistral-7B & 234 & 35 & 8 & 59 & 21 & 148 & 3 & 4 & 27 & 20 & 261 & 50 & 2 & 32 \\
 & 4 & Phi-3-medium-instruct & 326 & 48 & 12 & 64 & 25 & 380 & 5 & 9 & 62 & 48 & 529 & 70 & 4 & 76 \\
 & 5 & BLOOM-7B1 & 220 & 33 & 8 & 44 & 16 & 255 & 3 & 5 & 38 & 29 & 261 & 47 & 2 & 46 \\
 & 6 & Falcon-40B & 600 & 109 & 26 & 144 & 46 & 722 & 9 & 19 & 155 & 70 & 752 & 154 & 9 & 157 \\
 & 7 & GPT-J-6B & 146 & 17 & 4 & 53 & 20 & 245 & 2 & 6 & 14 & 26 & 154 & 40 & 3 & 53 \\
 & 8 & OPT-66B & 765 & 139 & 33 & 185 & 63 & 971 & 12 & 27 & 179 & 93 & 993 & 196 & 12 & 217 \\
 & 9 & Vicuna-13B & 314 & 47 & 11 & 63 & 24 & 363 & 5 & 8 & 61 & 46 & 502 & 67 & 4 & 74 \\
 & 10 & Vicuna-7B & 146 & 17 & 4 & 53 & 20 & 246 & 2 & 6 & 14 & 26 & 155 & 65 & 3 & 53 \\
 & 11 & BioMistral-7B & 235 & 35 & 9 & 49 & 17 & 269 & 3 & 5 & 43 & 32 &  272 & 49 & 2 & 48 \\
 & 12 & Medalpaca-7B & 192 & 24 & 5 & 39 & 14 & 98 & 2 & 2 & 17 & 13 & 172 & 53 & 1 & 21 \\
 & 13 & Vigogne-13B & 322 & 49 & 11 & 65 & 24 & 245 & 5 & 6 & 44 & 33 & 361 & 68 & 3 & 66 \\
\midrule
\multirow{16}{*}{\rotatebox[origin=c]{90}{Masked}} & 11 & mBERT & 14 & 4 & <1 & 2 & <1 & 15 & 1 & <1 & 1 & 1 & 13 & 2 & <1 & 2 \\
 & 12 & XLM-R-large & 14 & 4 & <1 & 2 & <1 & 15 & 1 & <1 & 1 & 1 & 13 & 2 & <1 & 2 \\
 & 13 & BERT-large &14 & 4 & <1 & 2 & <1 & - & - & - & - & - & - & - & - & - \\
 & 14 & RoBERTa-large &14 & 4 & <1 & 2 & <1 & - & - & - & - & - & - & - & - & - \\
 & 15 & Bio\_ClinicalBERT &14 & 4 & <1 & 2 & <1 & - & - & - & - & - & - & - & - & - \\
 & 16 & ClinicalBERT &14 & 4 & <1 & 2 & <1 & - & - & - & - & - & - & - & - & - \\
 & 17 & MedBERT &14 & 4 & <1 & 2 & <1 & - & - & - & - & - & - & - & - & - \\
 & 18 & CamemBERT-large &- & - & - & - & - & 15 & 1 & <1 & 1 & 1 & - & -& - & - \\
 & 19 & FlauBERT-large &- & - & - & - & - & 15 & 1 & <1 & 1 & 1 & - & -& - & - \\
 & 20 & DrBERT-4GB &- & - & - & - & - & 17 & 1 & <1 & 1 & 1 & - & -& - & - \\
 & 21 & CamemBERT-bio&- & - & - & - & - & 15 & 1 & <1 & 1 & 1 & - & -& - & - \\
 & 22 & BETO &- & - & - & - & -& - & - & - & - & - & 13 & 2 & <1 & 2 \\
 & 23 & PatanaBERT &- & - & - & - & -& - & - & - & - & - & 13 & 2 & <1 & 2 \\
 & 24 & TulioBERT &- & - & - & - & -& - & - & - & - & - & 13 & 2 & <1 & 2 \\
 & 25 & BSC-BioEHR &- & - & - & - & -& - & - & - & - & - & 13 & 2 & <1 & 2 \\
 & 26 & BSC-Bio &- & - & - & - & -& - & - & - & - & - & 13 & 2 & <1 & 2 \\
\midrule
\midrule
\multicolumn{17}{l}{\textit{Masked fully-supervised (skyline)}} \\
\midrule
 & & RoBERTa-large &14 & 4 & <1 & 2 & <1 & - & - & - & - & - & - & - & - & - \\
 & & CamemBERT-large &- & - & - & - & - & 15 & 1 & <1 & 1 & 1 & - & -& - & - \\
 & & BETO &- & - & - & - & -& - & - & - & - & - & 13 & 2 & <1 & 2 \\
\bottomrule
\end{tabular}}}
\caption{This table presents the carbon emissions (in g) of the inference on the test set of each model over each dataset.}
\label{tab:impact_test}
\end{table*}

\end{document}